\def\year{2022}\relax
\documentclass[letterpaper]{article} 
\usepackage{aaai22}  
\usepackage{times}  
\usepackage{helvet}  
\usepackage{courier}  
\usepackage[hyphens]{url}  
\usepackage{graphicx} 
\usepackage{natbib}  
\usepackage{caption} 
\DeclareCaptionStyle{ruled}{labelfont=normalfont,labelsep=colon,strut=off} 
\usepackage{algorithm}
\usepackage{algorithmic}
\usepackage{graphicx}

\usepackage{newfloat}
\usepackage{listings}
\usepackage{subfigure}
\usepackage{amsmath}
\usepackage{xcolor}

\DeclareMathOperator*{\argmin}{arg\,min}
\floatstyle{ruled}
\newfloat{listing}{tb}{lst}{}
\floatname{listing}{Listing}

\nocopyright

\title{Partial to Whole Knowledge Distillation:~Progressive Distilling Decomposed Knowledge Boosts Student Better}

\author{Xuanyang Zhang \hspace{20pt} Xiangyu Zhang \hspace{20pt} Jian Sun} 
\affiliations{MEGVII Technology \\
{\tt\small \{zhangxuanyang,zhangxiangyu,sunjian\}@megvii.com}
}

\begin{document}

\maketitle
\renewcommand{\thefootnote}{\fnsymbol{footnote}}
\footnotetext{This work is supported by The National Key Research and Development Program of China (No.2017YFA0700800) and Beijing Academy of Artificial Intelligence (BAAI).} 
\begin{abstract}
Knowledge distillation field delicately designs various types of knowledge to shrink the performance gap between compact student and large-scale teacher. These existing distillation approaches simply focus on the improvement of \textit{knowledge quality}, but ignore the significant influence of \textit{knowledge quantity} on the distillation procedure. Opposed to the conventional distillation approaches, which extract knowledge from a fixed teacher computation graph, this paper explores a non-negligible research direction from a novel perspective of \textit{knowledge quantity} to further improve the efficacy of knowledge distillation. We introduce a new concept of knowledge decomposition, and further put forward the \textbf{P}artial to \textbf{W}hole \textbf{K}nowledge \textbf{D}istillation~(\textbf{PWKD}) paradigm. Specifically, we reconstruct teacher into weight-sharing sub-networks with same depth but increasing channel width, and train sub-networks jointly to obtain decomposed knowledge~(sub-networks with more channels represent more knowledge). Then, student extract partial to whole knowledge from the pre-trained teacher within multiple training stages where cyclic learning rate is leveraged to accelerate convergence. Generally, \textbf{PWKD} can be regarded as a plugin to be compatible with existing offline knowledge distillation approaches. To verify the effectiveness of \textbf{PWKD}, we conduct experiments on two benchmark datasets:~CIFAR-100 and ImageNet, and comprehensive evaluation results reveal that \textbf{PWKD} consistently improve existing knowledge distillation approaches without bells and whistles.
\end{abstract}
\section{Introduction}

To make deep neural networks~(DNNs) practically applied, the demand for developing compact DNNs came into being. Shrinking the performance gap between compact models and large-scale models is of paramount importance. Knowledge distillation~(KD)~\cite{hinton2015distilling} is one of the representative schemes to develop compact models by distilling knowledge from teacher~(large-scale models) to student~(compact models). Up to now, there have derived many different paradigms, such as offline distillation~\cite{hinton2015distilling, romero2014fitnets}, online distillation~\cite{zhang2018deep, anil2018large} etc. More importantly, these derivative knowledge distillation paradigms have been widely used in computer vision~\cite{peng2019few,wu2020learning, zhang2020knowledge,li2017mimicking,mullapudi2019online} and natural language processing~(NLP) tasks~\cite{kim2016sequence, zhou2019understanding,tan2019multilingual, sun2019patient}. 
KD community attributes the success of knowledge distillation to \textit{dark knowledge}~\cite{hinton2015distilling} and hence mainly focuses on boosting student's performance from the perspective of \textit{knowledge quality}. To capture the nature of teacher representation, various types of knowledge, such as response-based knowledge~\cite{hinton2015distilling, tian2019contrastive}, feature-based knowledge~\cite{romero2014fitnets, zagoruyko2016paying,wang2020exclusivity}, relation-based knowledge~\cite{yim2017gift,tung2019similarity,passalis2020heterogeneous} and etc are designed to improve \textit{knowledge quality}.

\begin{figure}
   \centering
   \includegraphics[scale=0.34]{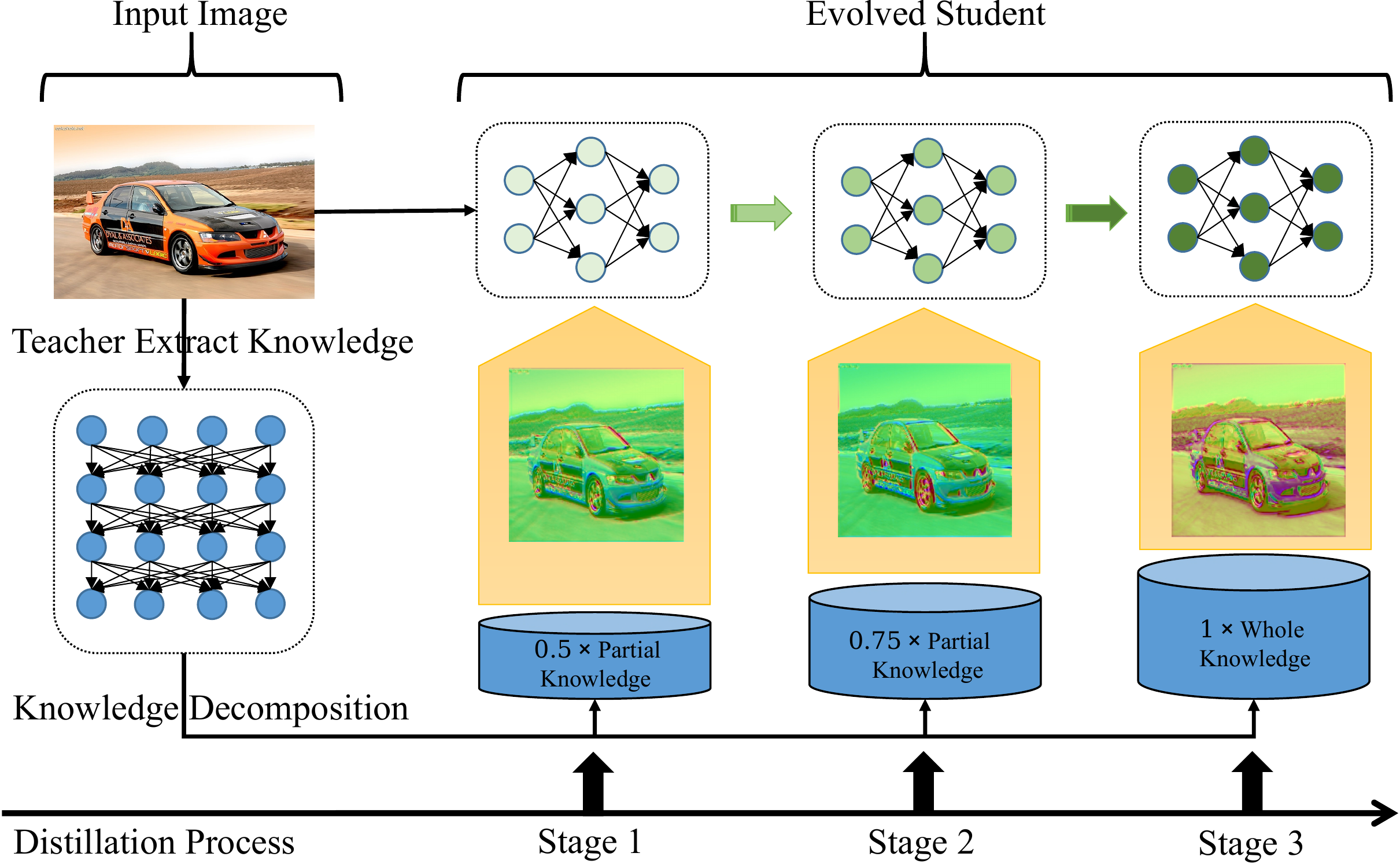}
   \vspace{-2.0em}
   \caption{The conceptual diagram of PWKD. Teacher decomposes the knowledge extracted from the input image into multiple pieces and then progressively transfers the increasing knowledge to student.}
   \label{fig.0}
\end{figure}

Despite the great success of KD, there are a few works to interpret what student benefits from \textit{dark knowledge}. \citet{cheng2020explaining} explains the working mechanism of knowledge distillation by quantifying the knowledge encoded in the intermediate layer, and they find out that knowledge distillation makes student learn more visual concepts. In this paper, we do not intend to further demystify \textit{dark knowledge}, but this interesting observation motivates us to rethink knowledge distillation schemes from the perspective of \textit{knowledge quantity} rather than \textit{knowledge quality}. Orange arrows in Figure~\ref{fig.0} represents the feature information corresponding to different knowledge quantity with one given knowledge type. \citet{cho2019efficacy} empirically finds that the teacher accuracy is a poor predictor of student performance, which implies that teacher consisting of more knowledge may not be a better teacher. TAKD~\cite{mirzadeh2020improved} introduces a teacher assistant with intermediate less knowledge to make the students learn better. Thus, we suppose that \textit{knowledge quantity} of teacher has a great influence on the efficacy of knowledge distillation. 

Based on the above observation, we survey most knowledge distillation literature whether it is offline or online paradigms. Then, we find out that teacher transfer the whole knowledge~(logits, intermediate feature maps, etc obtained with the whole feed-forward computation graph) to student at once, which is counter-intuitive in the view of human beings.  Imagining that a human teacher imparts all of what he has learned throughout his life to student, is it easy for student to accept it, especially in the early learning phases with poor cognitive ability? Hinton describes GLOM~\cite{hinton2021represent} to parse an image into a part-whole hierarchy, which motivates us to model the knowledge quantity of teacher into the partial-whole paradigm. We further make an intuitive hypothesis that progressively distilling knowledge from teacher with a partial-whole paradigm boosts student better~(named \textit{partial-whole hypothesis} for short).

To verify \textit{partial-whole hypothesis}, we propose a partial to whole knowledge distillation~(PWKD) paradigm as Figure~\ref{fig.0} shows. We define a new setup of \textit{knowledge decomposition} to parse knowledge of teacher into partial-whole knowledge. Different from GLOM~\cite{hinton2021represent} parsing image into independent partial representation, our goal is much simpler and aims to decompose knowledge representation into monotonically increasing fragments. Specifically, we reconstruct teacher into multiple weight-shared sub-networks with the same layer depth but different channel width, which preserves that knowledge quantity positively correlates with channel width~(\textit{e.g.}, knowledge of 1.0$\times$ teacher is more than the one of 0.5$\times$ teacher). Then, all sub-networks are trained jointly to obtain decomposed knowledge pieces. In the distillation process, we split the distillation into multiple training stages, in which student extract different knowledge fragments. To converge to a local minimum in each stage, cyclical learning rate scheduler~\cite{smith2017cyclical} is leveraged to student to accelerate convergence in the distillation process.

We empirically evaluate PWKD on two benchmark datasets:~CIFAR-100 and ImageNet, and PWKD is used as a general plugin compatible with nine mainstream offline distillation approaches~(\textit{e.g.}, FitNet~\cite{romero2014fitnets}, AT~\cite{sun2019patient}, CRD~\cite{tian2019contrastive} and etc). Comprehensive experiments show that knowledge distillation plugged with PWKD consistently improves performance with a substantial margin across various datasets, distillation approaches, and teacher-student pairs.

We summary the main contributions of this paper as follows:
\begin{enumerate}
    \item{For the first time, we analyze knowledge distillation from a new perspective of teacher \textit{knowledge quantity} instead of obsessing over \textit{knowledge quality}. }
    \item{We present the \textit{partial-whole hypothesis} and put forward a PWKD scheme to verify this hypothesis. Without loss of generality, PWKD is compatible with almost all offline distillation approaches.}
    \item{Comprehensive experiment results demonstrate that PWKD is such a simple yet effective distillation paradigm, which consistently improves distillation baseline with a substantial margin. The ablation study also strongly supports the intuition of this paper. We believe these results will further inspire a new understanding of knowledge distillation.}
\end{enumerate}


\section{Related Work}
\paragraph{Knowledge Distillation~(KD).} KD~\cite{hinton2015distilling} has developed to present and mainly consists of two distillation schemes:~offline distillation~\cite{hinton2015distilling, romero2014fitnets, huang2017like, heo2019knowledge, passalis2018learning} and online distillation~(self-distillation is regarded as a special online distillation)~\cite{zhang2018deep, anil2018large, zhang2019your, hou2019learning}. Offline distillation is the earliest distillation scheme and it includes two sequential training phases:~firstly, training a teacher model before distillation; secondly, the pre-trained teacher is used to guide the optimization of student. However, teacher with higher performance is not always available or training teacher costs large computation. Thus online distillation came into being. Online distillation has only one training phase, during which both teacher and student are optimized from scratch and updated simultaneously. To obtain decomposed knowledge, teacher must be pre-trained, and hence we mainly focus on the offline distillation scheme.

For offline distillation, it is inevitable to consider two questions:~(\romannumeral1)~what knowledge to distill? and (\romannumeral2) how to distill knowledge? To answer the first question, there are three categories of knowledge from teacher:~output logits~\cite{hinton2015distilling,chen2017learning,zhang2019fast}, intermediate feature map~\cite{romero2014fitnets,huang2017like,passalis2018learning} and relation-based knowledge~\cite{yim2017gift,lee2019graph}. Although the three types of knowledge represent different information, they share the same pattern that knowledge is obtained from the whole computation graph of teacher. There are few literatures that try to better distill knowledge. Opposed to the all-in distillation paradigm, we follow a similar rationale of curriculum distillation and propose a novel partial to whole knowledge distillation~(PWKD) paradigm to answer the second question in this paper.

\paragraph{Adaptive Neural Networks~(ANNs).} ANNs have attracted increasing attention because of their advantages in computation efficiency, representation power and etc~\cite{han2021dynamic}. Opposed to static ones, ANNs dynamically adjust structures according to input samples~\cite{wang2018skipnet,huang2017multi} or constrained computation budgets~\cite{yu2018slimmable,wang2020resolution}. More specifically, ANNs adjust structures from the perspective of depth~\cite{wang2018skipnet,huang2017multi}, channel width~\cite{yu2018slimmable, gao2018dynamic}, spatial resolution~\cite{wang2020resolution,yang2020resolution} or dynamic routing~\cite{li2020learning}. To answer the second question above~(how to distill knowledge?), we instantiate knowledge decomposition with the adjusted attribute of ANNs in the channel dimension. Different from most ANNs, which aim for the trade-off between accuracy and computation efficiency, we reconstruct static teacher into ANNs to obtain decomposed knowledge. Sub-networks preserve both coarse-grained and fine-grained features, and the relationship between the amount of knowledge and the number of sub-network channel widths can be clearly defined. 

\section{Methodology}
\label{methodology}
We clarify the goal of this paper is to build a new distillation setup:~partial to whole knowledge distillation~(PWKD) framework. To achieve this goal, we introduce the concept of \textit{knowledge decomposition}. To our best knowledge, it is the first time to involve \textit{knowledge decomposition} in the field of knowledge distillation. We reconstruct teacher into weight-sharing sub-networks with increasing channel widths and train these networks jointly to obtain decomposed knowledge. Then, we specify the interaction between student and teacher. At last, we put forward the overall PWKD framework that student progressively extracts partial to whole knowledge from teacher.

\subsection{Knowledge Decomposition}
\label{3.1}
Before introducing \textit{knowledge decomposition}, we first define the knowledge contained in a neural network. To keep the definition simple, we take the classical image classification as an example. Given a training dataset consisting of image and label tuples $(x,y) \in \mathcal{X} \times \mathcal{Y}$, a neural network $f(x, \mathbf{W})$ with learnable weights $\mathbf{W}$ is build to fit the mapping: $\mathcal{X} \mapsto \mathcal{Y}$. Cross entropy is used as the loss function~($\mathcal{L}$) and $\mathbf{W}^{\star}$ is achieved with Stochastic Gradient Descent~(SGD) optimization:

\begin{equation}
    \mathbf{W}^{\star} = \mathop{\argmin}_{\mathbf{W}} \mathcal{L}(f(x,\mathbf{W}), y).
\label{eq.1}
\end{equation}

Then, the well-optimized neural network $f(x, \mathbf{W}^{\star})$ is used for inference. Given an input $x$, $f(x, \mathbf{W}^{\star})$ propagates forward and the output logits, intermediate feature maps, and relationships between layers or samples can be defined as the knowledge $\Phi$.

\begin{figure}[htbp]
   \centering
   \includegraphics[scale=0.5]{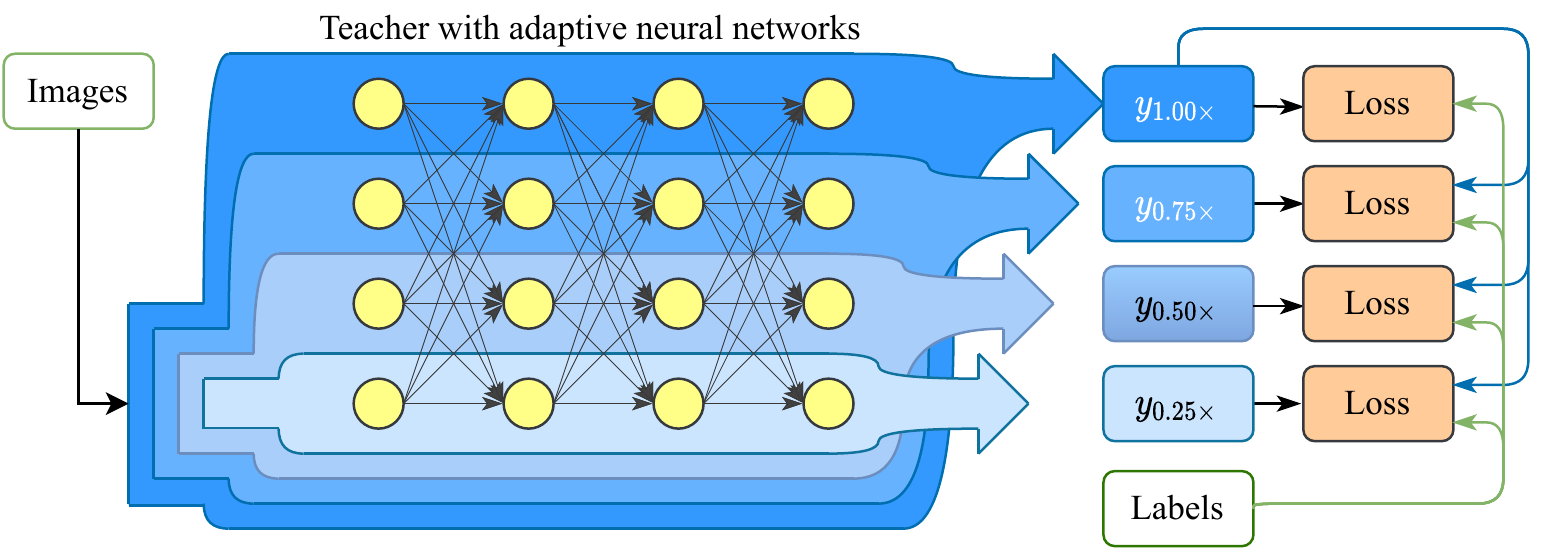}
   \vspace{-1.0em}
   \caption{The framework of knowledge decomposition. Teacher is split into four sub-networks with same layer depth but different channel width~(scale 0.25$\times$, 0.5$\times$, 0.75$\times$ and 1.0$\times$ channel width of full model). The knowledge quantity of four sub-networks is $\Phi_{0.25\times}$, $\Phi_{0.5\times}$, $\Phi_{0.75\times}$ and  $\Phi_{1.0\times}$ respectively.  }
   \label{fig.1}
\end{figure}

Because $\Phi$ is obtained with the whole network computation graph before the knowledge output node, we further define $\Phi$ as the \textit{'whole knowledge'}. In the knowledge distillation field, the conventional distillation approaches transfer \textit{'whole knowledge'} directly to student in the whole student optimization procedure~(name \textit{all-in scheme} for short). We argue that the \textit{all-in scheme} is much simpler but counter-intuitive. Knowledge distillation should obey the rule of curriculum learning from easy to difficult, just like humans. Intuitively, we can quantify the difficulty of learning from the perspective of \textit{knowledge quantity}. Thus, it is necessary to decompose teacher's knowledge before knowledge distillation.

Adaptive neural networks~(ANNs) adapt structures to the input or computation constrain, which implies that network representation can be split into multiple parts. Inspired by the partial activation property of ANNs, we can also decompose the knowledge of teacher by reconstructing teacher into multiple sub-networks. To clearly quantify knowledge, we need to explicitly define the correspondence between the sub-networks and the amount of knowledge. In general, model representation ability and model computational complexity are positively correlated. Therefore, an intuitive way to reconstruct teacher is to divide teacher into multiple sub-networks with the same depth but different channel widths and parameters are shared between different sub-networks. As Figure~\ref{fig.1} shows, we decompose the knowledge of vanilla teacher in four channel width groups~($0.25\times$, $0.5\times$, $0.75\times$, and $1.0\times$ channel width in all layers of the vanilla teacher). With bigger channel width, the sub-network has stronger prediction performance, which means that the sub-network contains more knowledge. We define the knowledge of sub-networks as \textit{'partial knowledge'}
After clarifying the correspondence between sub-networks and knowledge quantity, we train these sub-networks to obtain representation knowledge. Due to the weight-sharing strategy, we involve switchable-BN and switchable-classifier~\cite{yu2018slimmable}~(each sub-network has private BN~\cite{ioffe2015batch} in each layer and classifier layer) into each sub-networks to prevent performance collapse. Then, we train 1.0$\times$ teacher with cross entropy loss~($\mathcal{L}_{cls}$):
\begin{equation}
    \mathcal{L} = \mathcal{L}_{cls}(f_{t}(x,\mathbf{W_{1.0\times}}), y),
\label{eq.2}
\end{equation}
and train the other sub-networks with both cross entropy loss~($\mathcal{L}_{cls}$) and Kullback-Leibler divergence loss~($\mathcal{L}_{kl}$):
\begin{equation}
\begin{split}
\mathcal{L} = & \alpha*\mathcal{L}_{cls}(f_{t}(x,\mathbf{W_{\rho\times}}), y) \\
& + (1-\alpha)*\mathcal{{L}}_{kl}(f_{t}(x,\mathbf{W}_{\rho\times}), f_{t}(x,\mathbf{W_{1.0\times}}), T),
\end{split}
\label{eq.3}
\end{equation}
where $\alpha$, $\rho$ and $T$ represent $\mathcal{L}_{cls}$ weighted factor, channel width scale and temperature factor. To make it simple, we set $\alpha=0.5$ and $T=1.0$ in this paper if there are no specific instructions. During each training iteration, all sub-networks propagate forward and backward successively. Then, the accumulated back-propagation gradients are applied to update teacher weights. After all, sub-networks are trained to converge, we can obtain the decomposed knowledge.

\begin{figure*}[htbp]
   \centering
   \includegraphics[scale=0.48]{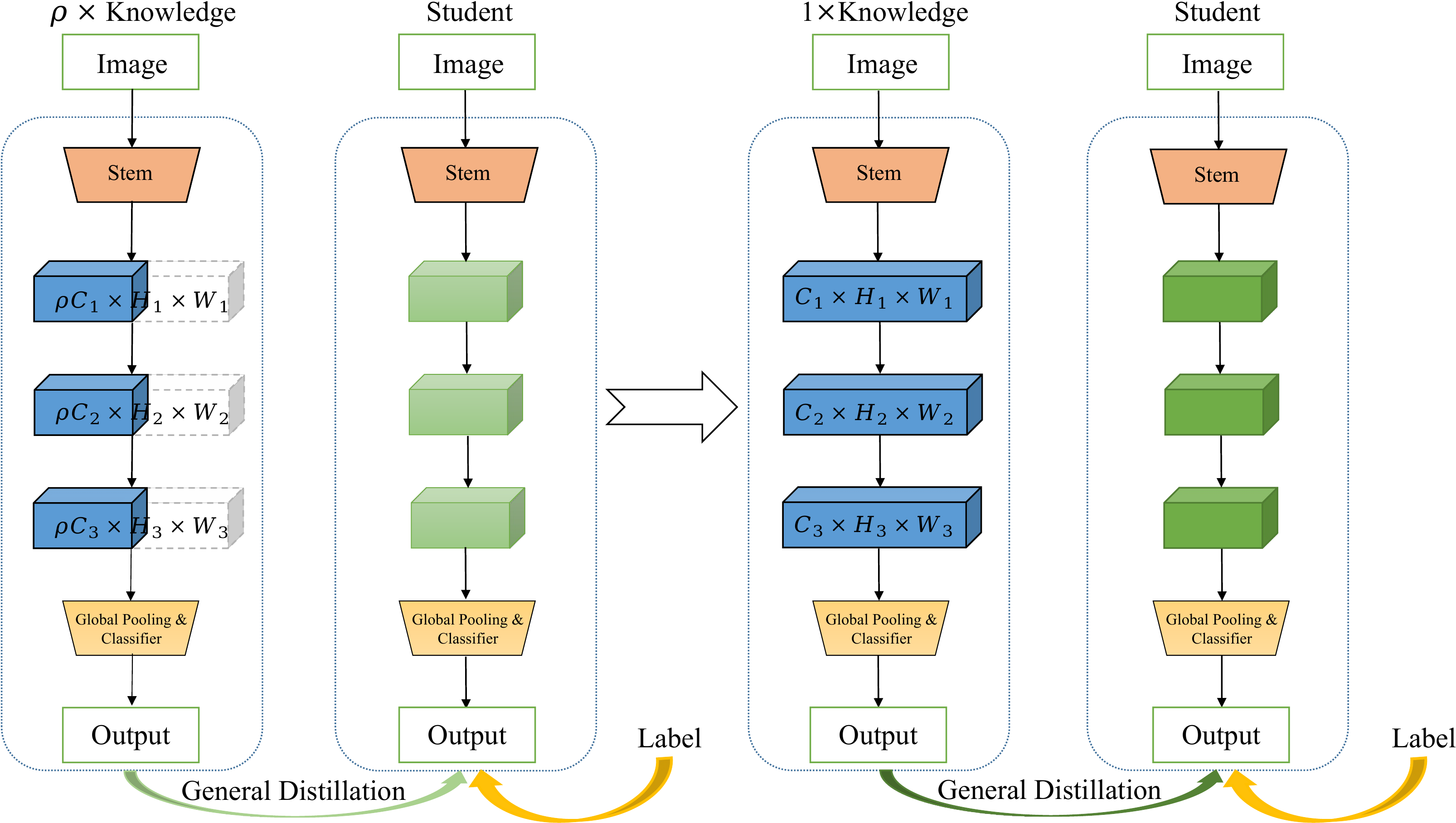}
   \hspace{5.0em}
   \raisebox{0.0em}{(a)~intermediate distillation stage with $\Phi_{\rho\times}$}
   \hspace{5.0em}
   \raisebox{0.0em}{(b)~final distillation stage with $\Phi_{1.0\times}$}
   \vspace{-1.0em}
   \caption{An illustration of partial to whole knowledge distillation~(PWKD) framework~(best viewed in color). The darker the green, the higher the student's prediction performance.}
   \label{fig.2}
\end{figure*}

\subsection{Distillation with Decomposed Knowledge}
\label{3.2}
Given the pre-trained teacher optimized with Eq.\ref{eq.2} and Eq.\ref{eq.3}, student can extract the decomposed knowledge $\Phi_{\rho\times}$ from the sub-networks with pre-defined channel width. Because the only difference between sub-networks and the full model is the channel width in each layer, the decomposed knowledge can be applied to almost all off-line distillation approaches. The whole loss function can be formulated as:
\begin{equation}
\begin{split}
\mathcal{L} = & \beta*\mathcal{L}_{cls}(f_{s}(x,\mathbf{W}_{s}), y) \\
              & + (1-\beta)*\mathcal{{L}}_{kd}(f_{s}(x,\mathbf{W}_{s}),f_{t}(x,\mathbf{W_{\rho\times}}), T),
\end{split}
\label{eq.4}
\end{equation}
where $\beta$ represents $\mathcal{L}_{cls}$ weighted factor in student training phase and $\mathcal{{L}}_{kd}$ stands for a general distillation loss, such as similarity loss~\cite{tung2019similarity}, contrastive loss~\cite{tian2019contrastive} and etc. The detailed distillation process can be described as Figure~\ref{fig.2}~(a).

\subsection{Overall Partial to Whole Distillation}
\label{3.3}
\paragraph{Partial to whole multiple stage distillation.}~Although students can distill decomposed knowledge arbitrarily, we argue that progressively knowledge extraction boosts student better. In the above subsection, we have explicitly quantified knowledge through model capability:~the more channels there are, the more knowledge the sub-network includes. Sequentially, we only need to divide the student distillation process into multiple stages and each stage distills knowledge gradually in ascending order. We keep the student training epochs as the vanilla distillation process and just split the total training epochs to each piece of knowledge with equal epochs. As Figure~\ref{fig.2} shows, teacher transfers knowledge to student from $\Phi_{\rho\times}$ to $\Phi_{1.0\times}$ progressively.

\paragraph{Cyclical learning rate scheduler.}~During most knowledge distillation approaches, they use the learning rate scheduler with a fixed value that monotonically decreases during the whole training procedure of students. That makes sense, because student just needs to mimic teacher with a piece of knowledge. However, in the PWKD framework, there are multiple pieces of knowledge and student need to mimic different pieces of knowledge in each distillation stage. The conventionally monotonous learning rate schedulers make student optimized with a bigger learning rate in early distillation stages and yet smaller learning rate in later distillation stages, which prevents students from fully absorbing knowledge and further limits the performance improvement.

To fully take advantage of each piece of knowledge, student must converge to a local minimum in each distillation stage. Cyclical learning rate~(CLR) scheduler can make model converge fast~\cite{smith2017cyclical} and we equip PWKD with CLR to converge to multiple local minimums. To make a fair comparison, we use the same training epochs as with other knowledge distillation approaches. Thus, each local minimum could be achieved with only $1/G$~($G$ represents the number of knowledge pieces) of the total training epochs. Specifically, CLR consists of multiple cycles with the same learning rate policy. In each cycle, the learning rate is constrained with a range and varies between the minimum and maximum boundary with a certain functional form. In this paper, we instantiate the cyclical function form as the triangular window, which linearly increases from the minimum to the maximum bound and then linearly decreases to the minimum with equal window size.   

\begin{figure*}[htbp]
   \centering
   \subfigure[Distilled from ResNet-20$\times$4]{
       \includegraphics[scale=0.28]{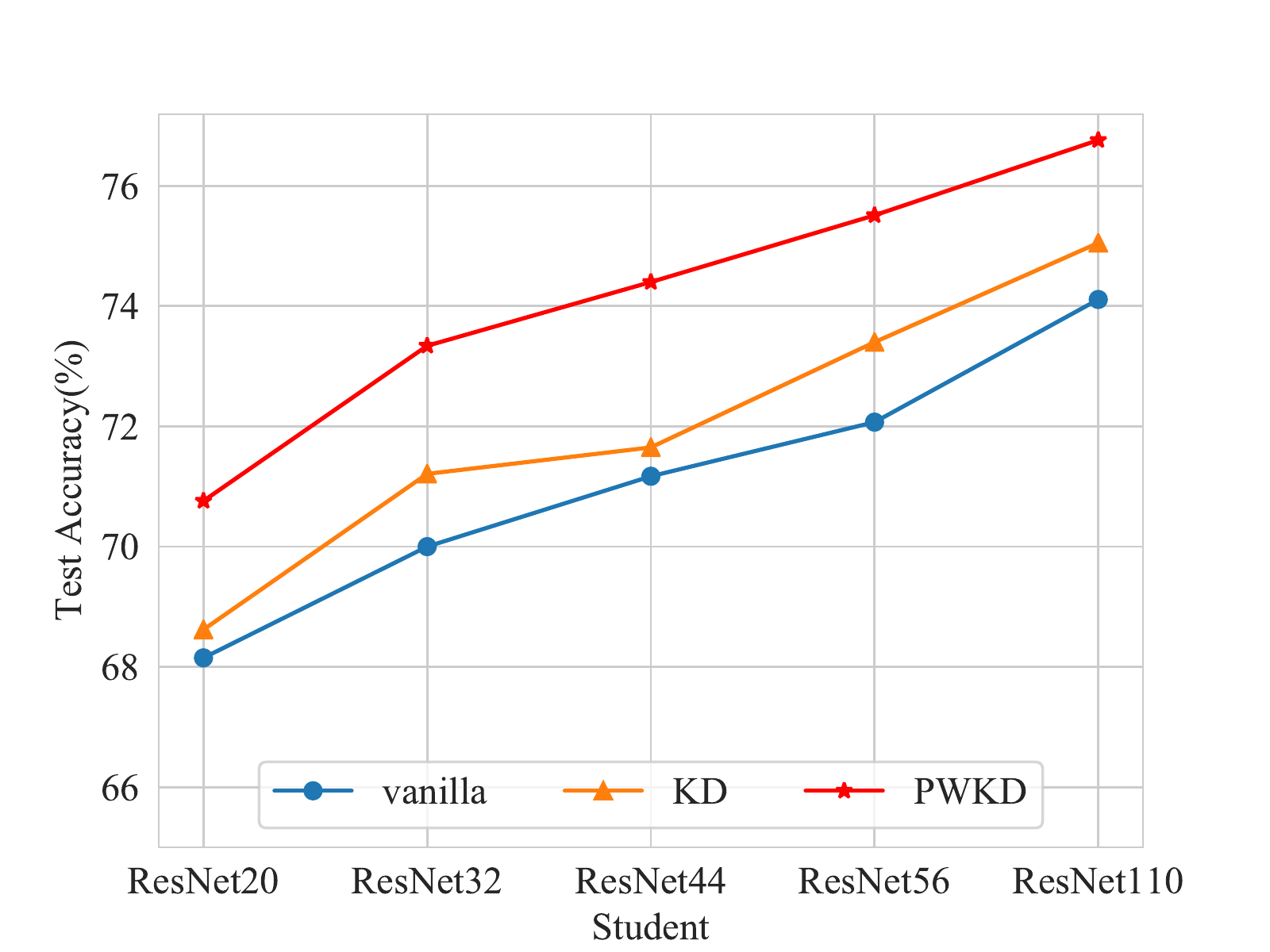}}
  \hspace{-5mm}
   \subfigure[Distilled from ResNet-32$\times$4]{
       \includegraphics[scale=0.28]{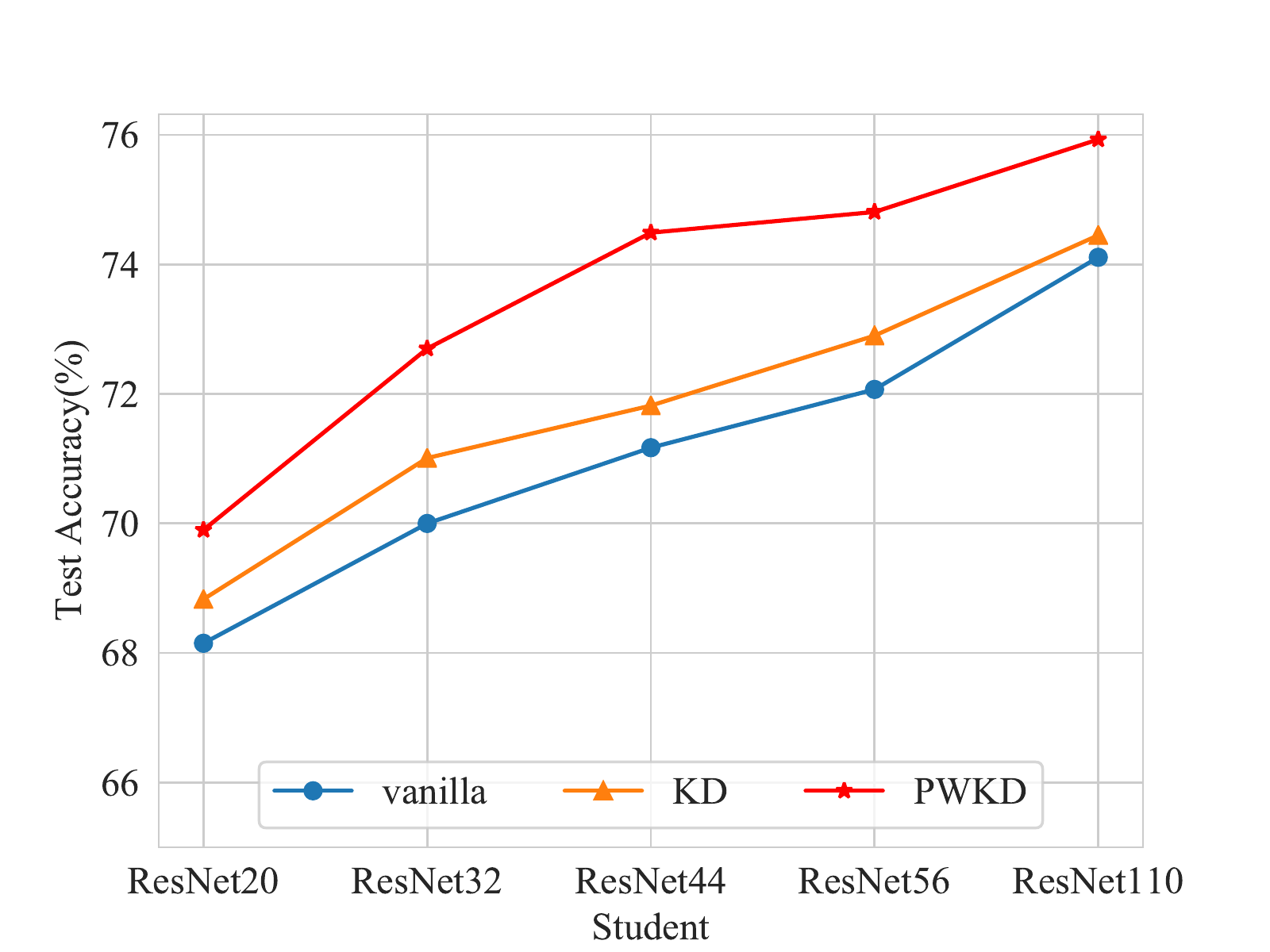}}
  \hspace{-5mm}
   \subfigure[Distilled from ResNet-44$\times$4]{
       \includegraphics[scale=0.28]{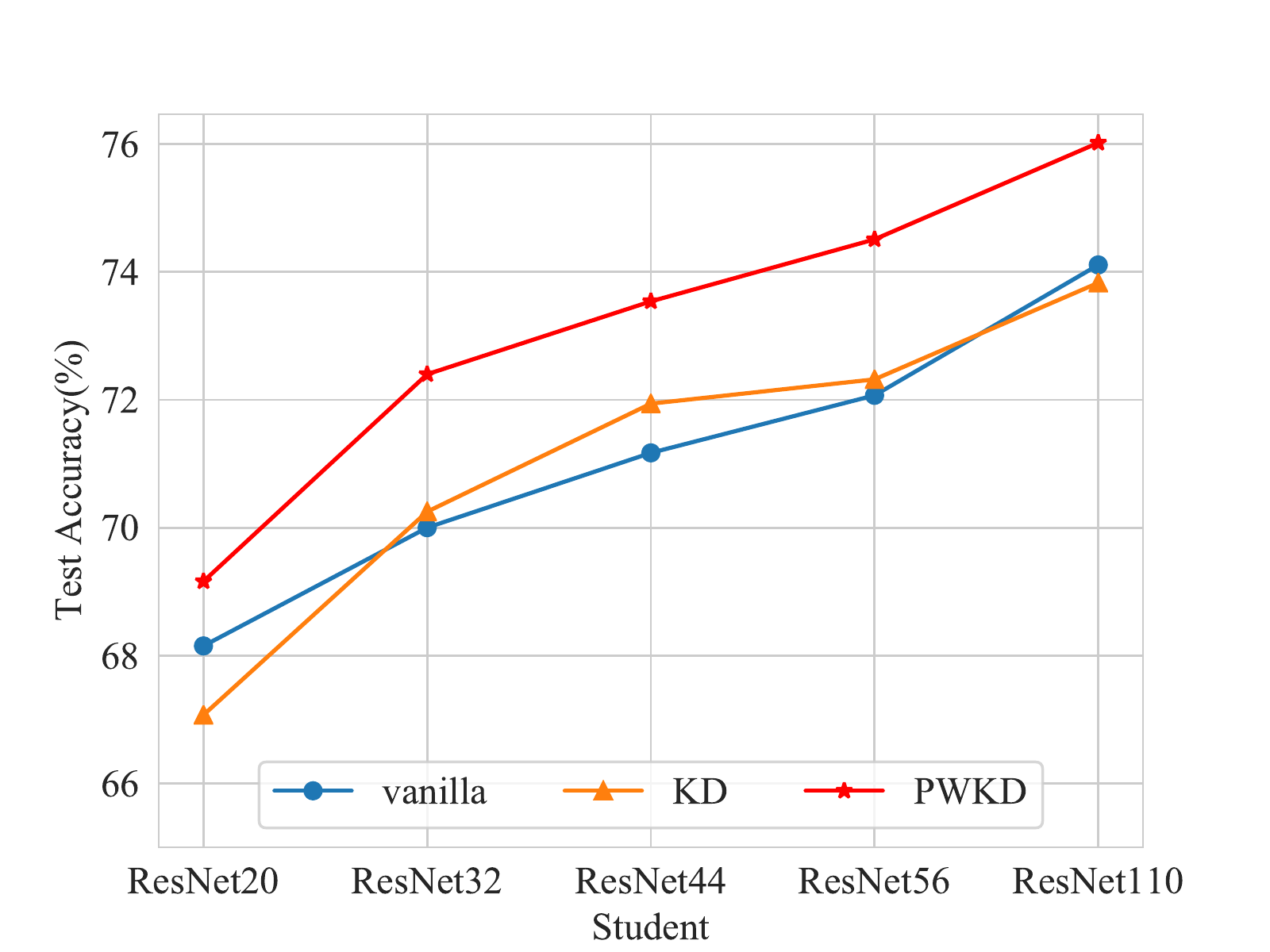}}
  \hspace{-5mm}
   \subfigure[Distilled from ResNet-56$\times$4]{
       \includegraphics[scale=0.28]{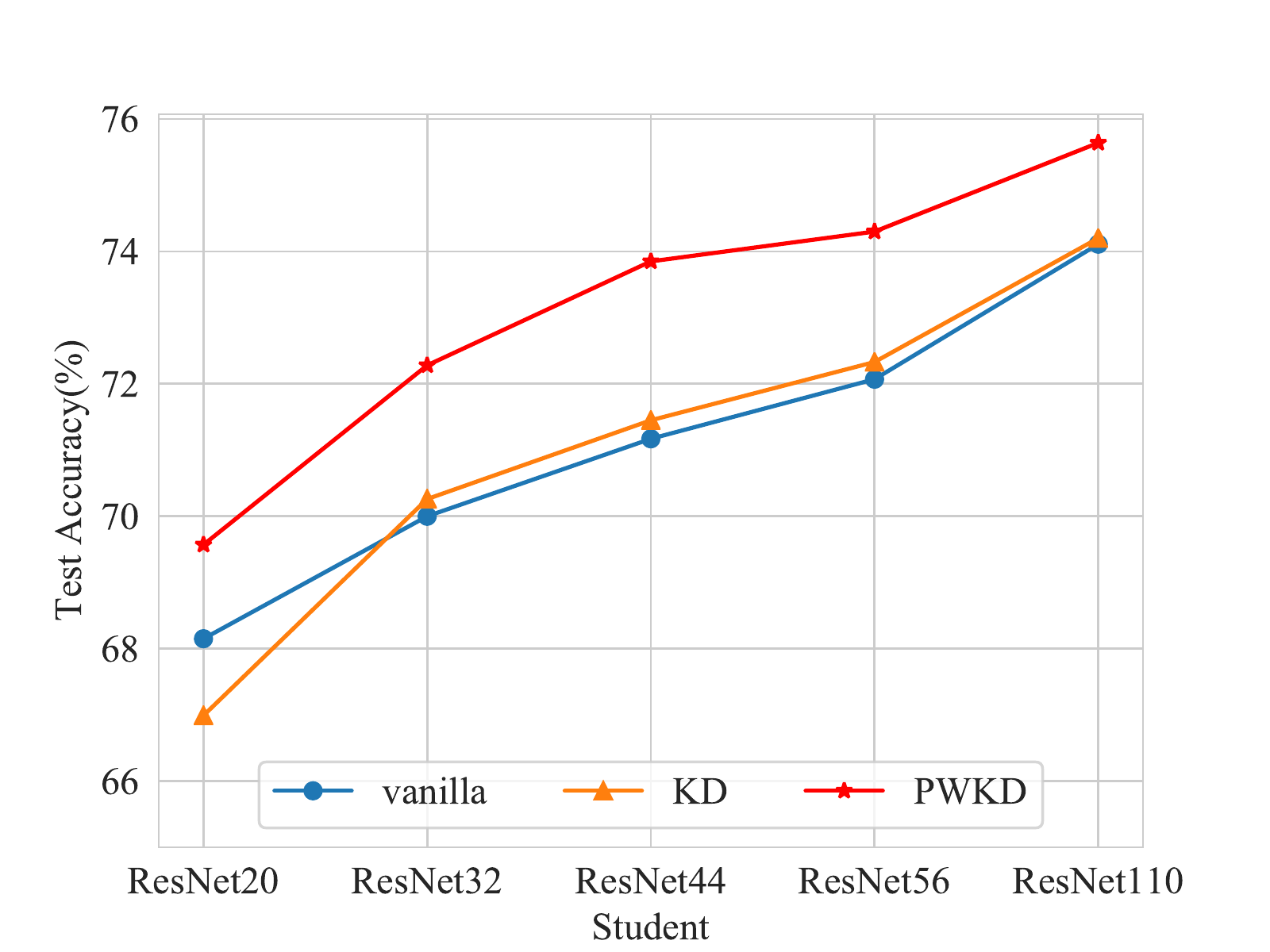}}
   \vspace{-1.5em}
   \caption{Comparison results between PWKD with two baselines:~student trained from scratch and distilled with KD across various teach-student pairs. Especially, PWKD still improve student substantial margin for ResNet-44$\times$4/ResNet-20 and ResNet-56$\times$4/ResNet-56, while KD fails to boost student in such case. }
   \label{fig.3}
\end{figure*}

\section{Experiments}
We evaluate partial to whole knowledge distillation~(PWKD) on CIFAR-100~\cite{krizhevsky2009learning} and ImageNet~\cite{deng2009imagenet}. We compare PWKD with standard KD on both datasets and to claim the generality, we extend PWKD to nine popular knowledge distillation methods on CIFAR-100. At last, we ablate the effect of PWKD and hyper-parameters.

\subsection{Main Results}
\label{main_result}
\subsubsection{Results on CIFAR-100}
In this section, we aim to verify the effectiveness of PWKD on CIFAR-100 dataset. Firstly, we reconstruct teacher into 4 weight-sharing sub-networks with channel width 0.25$\times$, 0.5$\times$, 0.75$\times$ and 1.0$\times$ respectively. Then, all sub-networks are trained jointly in each iteration to obtained knowledge fragments. Specifically, we set ResNet-20$\times$4, ResNet-32$\times$4, ResNet-44$\times$4 and ResNet-56$\times$4 as teacher~($\times4$ means that channel width of each layer expands 4 times), and the test accuracy of sub-networks are reported in Table~\ref{tab.1}. For each teacher, test accuracy of sub-networks increase monotonically with channel with, which means knowledge quantity is positively correlated with channel width. Further compared with test accuracy of teacher without reconstruction, we observe that all sub-networks has inferior performance even for sub-networks with channel width 1.0$\times$. This phenomenon may be caused by conflicted optimization goals of sub-networks in weight-sharing strategy.

With the pre-trained sub-networks, we split the distillation process into four stages, which equals to the number of sub-networks. In each distillation stage, student distill from one sub-network, and sub-networks are applied to each stage in the order of channel width from 0.25$\times$ to 1.0$\times$. Besides, the cyclical learning rate also consist of four cycles, each distillation state share the same learning rate decay policy to converge to local minimum.

We apply the above four teachers to five students:~ResNet-20, ResNet-32, ResNet-44, ResNet-56 and ResNet-110. We define test accuracy of student trained from scratch and distilled from teacher with KD~\cite{hinton2015distilling} as baselines. We sum up the evaluation results in Figure~\ref{fig.3} with four sub-figures. As Figure~\ref{fig.3} shows, PWKD consistently outperforms the two baselines for all teacher-student pairs. An interesting observation is that in Figure~\ref{fig.3}~(c) and Figure~\ref{fig.3}~(d), ResNet-20 with KD has inferior performance~(drop by \textbf{1.08\%} and \textbf{1.16\%}) than vanilla performance~(this phenomenon is interpreted as teacher-student gap in~\citet{mirzadeh2020improved}), but PWKD can still improve ResNet-20 distilled from ResNet-44$\times$4 and ResNet-56$\times$4 with \textbf{1.01\%} and \textbf{1.42\%} respectively. To be most important, as Table~\ref{tab.1} shows, all sub-networks in PWKD under-perform vanilla teacher, but distill various students outperform the ones distilled from vanilla teacher. These promising results demonstrate that PWKD boosts students better. 

\begin{table}[htbp]
   \scriptsize
   \begin{center}
   \begin{tabular}{c|c|c|c|c|c}
   \hline
   Model & 0.25$\times$ & 0.5$\times$ & 0.75$\times$ &  1.0$\times$ & vanilla \\
   \hline\hline
   ResNet-20$\times$4 & 68.88 & 74.67 & 76.59 & 77.43 & 78.58 \\
   \hline
   ResNet-32$\times$4 & 71.07 & 76.19 & 78.09 & 78.69 & 79.63 \\
   \hline
   ResNet-44$\times$4 & 71.81 & 76.66 & 78.08 & 78.65 & 79.14 \\
   \hline
   ResNet-56$\times$4 & 72.62 & 76.51 & 78.07 & 78.13 & 79.47 \\
   \hline
   VGG13 & 65.95 & 71.64 & 73.24 & 74.13 & 75.83  \\
   \hline
   \end{tabular}
   \end{center}   
   \vspace{-2em}
   \caption{Results of sub-networks on CIFAR-100. Each network is reconstructed into 4 sub-networks with channel width 0.25$\times$, 0.5$\times$, 0.75$\times$ and 1$\times$ respectively. Knowledge quantity of sub-networks increase monotonically with channel with, and all sub-networks consistently have inferior performance compared with vanilla teacher.} 
   \label{tab.1}
\end{table}

\subsubsection{Extend to various distillation approaches}

\begin{table*}[ht]
   \centering 
   \scriptsize
    
   \begin{center}
   \begin{tabular}{c|c|c|c|c|c|c|c|c|c|c|c|c} 
   \hline 
   Teacher/student pair &  \multicolumn{3}{c}{ResNet-20$\times$4/ResNet-20~(\%)} & \multicolumn{3}{|c}{ResNet-32$\times$4/ResNet-32~(\%)} & \multicolumn{3}{|c}{ResNet-44$\times$4/ResNet-44~(\%)} & \multicolumn{3}{|c}{ResNet-56$\times$4/ResNet-56~(\%)} \\
   \hline
   Distillation method & Vanilla & PWKD & $\Delta$ & Vanilla & PWKD & $\Delta$ & Vanilla & PWKD & $\Delta$ & Vanilla & PWKD & $\Delta$ \\
   \hline\hline
    Teacher accuracy & 78.58 & 77.43 & - & 79.63 & 78.69 & - & 79.14 & 78.65  & - & 79.47 & 78.13 & - \\
   Student w/o distillation & 68.15 & 68.15 & - &70.00 &70.00 & - &71.17 &71.17 &- &72.07 &72.07 & - \\
   \hline
   \hline
   KD & 68.62 & 70.76 & \textbf{+2.14} &71.01 & 72.70 & \textbf{+1.69} & 71.94 & 73.54 & \textbf{+1.60}  &72.33 & 74.30 & \textbf{+1.97} \\
   \hline
   FitNet & 67.84 & 69.73 & \textbf{+1.89} & 70.34 & 70.91 & \textbf{+0.57} & 71.28 & 72.40 & \textbf{+1.12} & 71.84 & 73.34 & \textbf{+1.50} \\
   \hline
   AT & 68.26 & 70.42 & \textbf{+2.16} & 70.46 & 72.42 & \textbf{+1.96} & 71.96 & 73.68 & \textbf{+1.72} & 72.10 & 74.45 & \textbf{+2.35} \\
   \hline
   SP & 69.72 & 70.75 & \textbf{+1.03} & 70.88 & 73.47 & \textbf{+2.59} & 72.00 & 74.53 & \textbf{+2.53} & 71.98 & 74.91 & \textbf{+2.93} \\
   \hline
   CC & 69.00 & 69.86 & \textbf{+0.86} & 70.38 & 72.13 & \textbf{+1.75} & 71.59 & 72.93 & \textbf{+1.34} & 71.60 & 73.08 & \textbf{+1.48} \\
   \hline
   VID & 69.16 & 70.37 & \textbf{+1.21} & 71.17 & 72.26 & \textbf{+1.09} & 71.52 & 72.77 & \textbf{+1.25} & 72.79 & 73.69 & \textbf{+0.90} \\
   \hline
   RKD & 68.84 & 70.08 & \textbf{+1.24} & 69.88 & 72.03 & \textbf{+2.15} & 71.42 & 73.59 & \textbf{+2.17} & 72.15 & 73.97 & \textbf{+1.82} \\
   \hline
   PKT & 69.70 & 70.50 & \textbf{+0.80} & 71.19 & 72.82 & \textbf{+1.63} & 72.54 & 73.92 & \textbf{+1.38} & 73.43 & 74.58 & \textbf{+1.15} \\
   \hline
   CRD & 69.90 & 71.59 & \textbf{+1.69} & 71.91 & 73.64 & \textbf{+1.73} & 72.32 & 74.41 & \textbf{+2.09} & 72.68 & 75.40 &  \textbf{+2.72} \\
   \hline
   \end{tabular}
   \end{center}
   \vspace{-2em}
   \caption{Comparison results between nine distillation methods and the corresponding ones plugged with PWKD. Across various homogeneous teacher-student pairs~(ResNet/ResNet), PWKD consistently outperforms the vanilla ones.} 
   \label{tab.2}
\end{table*}

\begin{table*}[ht]
   \scriptsize
   \begin{center}
   \begin{tabular}{c|c|c|c|c|c|c|c|c|c|c|c|c} 
   \hline 
   Teacher/student pair&  \multicolumn{3}{c}{ResNet-20$\times$4/ShuffleNet-V1~(\%)} & \multicolumn{3}{|c}{ResNet-20$\times$4/MobileNet-V2~(\%)} & \multicolumn{3}{|c}{VGG13/ShuffleNet-V1~(\%)} & \multicolumn{3}{|c}{VGG13/ResNet-20~(\%)} \\
   \hline
   Distillation method & Vanilla & PWKD & $\Delta$ & Vanilla & PWKD & $\Delta$ & Vanilla & PWKD & $\Delta$ & Vanilla & PWKD & $\Delta$ \\
   \hline\hline
    Teacher accuracy & 78.58 & 77.43 & -  & 78.58 & 77.43 & -  & 75.83 & 74.13  & - & 75.83  & 74.13 & - \\
   Student w/o distillation & 71.06 & 71.06 & - & 62.20 & 62.20 & - & 71.06 & 71.06 & - & 68.15 & 68.15 & - \\
   \hline
   \hline
   KD & 73.80 & 75.45 & \textbf{+1.65} &66.78 & 67.72 & \textbf{+0.94} & 73.18 & 74.30 & \textbf{+1.12}  & 69.08 & 69.75 & \textbf{+0.67} \\
   \hline
   FitNet & 72.84 & 74.05 & \textbf{+1.21} & 63.50 & 65.89 & \textbf{+2.39} & 71.06 & 72.88 & \textbf{+1.82} & 67.32 & 69.85 & \textbf{+1.53} \\
   \hline
   AT & 70.77 & 73.43 & \textbf{+2.66} & 60.70 & 63.73 & \textbf{+3.03} & 72.23 & 72.48 & \textbf{+0.25} & 66.04 & 67.35 & \textbf{+1.31} \\
   \hline
   SP & 74.79 & 75.18 & \textbf{+0.39} & 64.79 & 66.68 & \textbf{+1.89} & 74.80 & 75.35 & \textbf{+0.55} & 68.91 & 69.85 & \textbf{+0.94} \\
   \hline
   CC & 71.07 & 72.03 & \textbf{+0.96} & 63.48 & 65.60 & \textbf{+2.12} & 70.62 & 72.57 & \textbf{+1.95} & 67.91 & 69.70 & \textbf{+1.79} \\
   \hline
   VID & 73.49 & 74.16 & \textbf{+0.67} & 63.34 & 65.68 & \textbf{+2.34} & 71.04 & 72.31 & \textbf{+1.27} & 69.13 & 69.39 & \textbf{+0.26} \\
   \hline
   RKD & 72.19 & 74.03 & \textbf{+1.84} & 62.00 & 66.71 & \textbf{+4.71} & 72.74 & 73.05 & \textbf{+0.31} & 68.77 & 69.67 & \textbf{+0.90} \\
   \hline
   PKT & 73.88 & 74.37 & \textbf{+0.49} & 65.13 & 67.05 & \textbf{+1.92} & 73.21 & 73.36 & \textbf{+0.15} & 68.93 & 70.16 & \textbf{+1.23} \\
   \hline
   CRD &74.09  & 75.70  &  \textbf{+1.61} & 67.32 & 69.43 &  \textbf{+2.11} & 73.84 & 75.70 & \textbf{+1.86} & 69.91 & 70.64 & \textbf{+0.73} \\
   \hline
   \end{tabular}
   \end{center}
   \vspace{-2em}
   \caption{Comparison results between nine distillation methods and their corresponding ones armed with PWKD. Across various heterogeneous teacher-student pairs~(ResNet/ShuffleNet, ResNet/MobileNet, VGGNet/ShuffleNet, and VGGNet/MobileNet), distillation methods plugged with PWKD consistently outperform the vanilla ones.}
   \label{tab.3}
\end{table*}

As described in the methodology, PWKD can be used as a general plugin to be applied to various knowledge distillation approaches. To support the generality, we extend PWKD to existing distillation approaches in this section from the perspectives of teacher-student architecture styles and distillation methods. For teacher-student architecture styles, we mainly consider teacher-student pairs with homogeneous architectures~(\textit{e.g.}, ResNet/ResNet~\cite{he2016deep}) and heterogeneous architectures~(\textit{e.g.}, ResNet/ShuffleNet~\cite{zhang2018shufflenet}, ResNet/MobileNet~\cite{sandler2018mobilenetv2}, VGGNet~\cite{simonyan2014very}/ShuffleNet and VGGNet/ResNet). For distillation methods, we introduce nine popular algorithms, consisting of KD~\cite{hinton2015distilling}, FitNet~\cite{romero2014fitnets}, AT~\cite{zagoruyko2016paying}, SP~\cite{tung2019similarity}, CC~\cite{peng2019correlation}, VID~\cite{ahn2019variational}, RKD~\cite{park2019relational}, PKT~\cite{passalis2018learning}, NST~\cite{huang2017like} and CRD~\cite{tian2019contrastive}. We keep most training setting the repository~\footnote{https://github.com/HobbitLong/RepDistiller} provided in CRD~\cite{tian2019contrastive}, except that we train both teacher and student 320 epochs, batch size 256 accompanied with 2$\times$ initial learning rate.

We report evaluation results of homogeneous and heterogeneous teacher-student pairs in Table~\ref{tab.2} and Table~\ref{tab.3} respectively. In each table, there are four different teacher-student pairs. As Table~\ref{tab.2} and Table~\ref{tab.3} shows, teacher plugged with PWKD still consistently boosts student better across two architecture styles and nine popular distillation methods. In Table~\ref{tab.2}, PWKD achieve a maximum gain \textbf{2.93\%}, and the average improvement is 1.67\%. Simultaneously, PWKD achieves a maximum gain \textbf{4.71\%}, and the average improvement is 1.43\% in Table~\ref{tab.3}. Thus, we conclude that PWKD is such general distillation paradigm, which is independent from architecture styles and knowledge categories.

\subsubsection{Results on ImageNet}
\begin{table}[h]
  \scriptsize
   \begin{center}
   \begin{tabular}{c|c|c|c|c|c}
   \hline
   Model & 0.25$\times$ & 0.5$\times$  & 0.75$\times$ & 1.0$\times$ & vanilla \\
   \hline\hline
   ResNet-50 & 64.70 & 72.40 & 75.48 & 76.91 & 75.47 \\
   \hline
   \end{tabular}
   \end{center}   
   \vspace{-2.0em}
   \caption{Top-1 accuracy results of sub-networks on ImageNet. ResNet-50 is reconstructed into 4 sub-networks with channel width 0.25$\times$, 0.5$\times$, 0.75$\times$ and 1$\times$ respectively. Knowledge quantity of sub-networks increase monotonically with channel.} 
   \label{tab.4}
\end{table}

\begin{table}[h]
  \scriptsize
  \begin{center}
  \begin{tabular}{c|c|c|c|c}
  \hline
  Teacher/student pair & student & vanilla KD & PWKD & $\Delta$ \\
  \hline\hline
  ResNet-50/ResNet10 & 62.53  & 62.84 & 63.69 & \textbf{+0.85} \\
  \hline
  ResNet-50/ResNet-18 & 69.36  & 70.16 & 70.84 & \textbf{+0.68}  \\
  \hline
  ResNet-50/ResNet-34 & 73.07  & 73.86 & 74.46 & \textbf{+0.60}\\
  \hline
  \end{tabular}
  \end{center}
  \vspace{-2em}
  \caption{Comparison results on ImageNet. We compare PWKD with two baselines: student trained from scratch and student distilled with vanilla KD. $
  \Delta$ represents the performance improvement of PWKD compared with vanilla KD.} 
  \label{tab.5}
\end{table}

We further evaluate PWKD on ImageNet. ResNet-50 is adopt as the teacher, and we reconstruct ResNet-50 into 4 sub-networks with also 50 layers but 0.25$\times$, 0.5$\times$, 0.75$\times$ and 1$\times$ channel width respectively. We train the four sub-networks jointly to obtain knowledge $\Phi_{0.25\times}$, $\Phi_{0.5\times}$, $\Phi_{0.75\times}$ and $\Phi_{1\times}$. Top-1 accuracy of sub-networks are reported in Table~\ref{tab.1}. There are something different from sub-networks results on CIFAR-100 that sub-network ResNet-50 with 1$\times$ channel width outperforms vanilla ResNet-50 trained from scratch independently.

When applying KD to student, we find default KD setting $\beta$=0.9 and T=6.0 can not improve student as observed in \cite{cho2019efficacy}. After comparative analysis with CIFAR-100, we guess this phenomenon is caused by inferior training accuracy and larger image category. Thus we try to assign more credit to ground truth labels may improve the efficacy of KD. Specifically, we set $\beta$ in Eq.4 to 0.1. As Table~\ref{tab.4} shows, vanilla KD improve student ResNet10, ResNet-18 and ResNet-34 0.32\%, 0.80\% and 0.79\%. Following this hyper-parameter setting, we plug vanilla KD with PWKD, and maximum and minimum boundary of CLR are 0.1 and 0.0001. The performance of students can be further improved by 0.85\%, 0.68\% and 0.60\% compared with vanilla KD methods.

\subsection{Ablation study}
\label{ablation_study}
\subsubsection{Effects of PWKD and cyclical scheduler}

Our proposed distillation scheme consists of PWKD and cyclical learning rate scheduler. We choose ResNet-20$\times$4/ResNet-20 as teacher/student pair to ablate the effect of PWKD and cyclical learning rate~(CLR) in this section. As Table~\ref{tab.6} shows, vanilla KD improves ResNet-20 with 0.47\%, but PWKD without cyclical learning rate scheduler obtains 2.29\% improvement compared with baseline. Further, we introduce cyclical learning rate scheduler to PWKD and PWKD improve ResNet-20 2.61\%, which imply convergence to multiple local minimums prompt student better.  
\begin{table}[htbp]
   \scriptsize
   \begin{center}
   \begin{tabular}{lcccc}
   \hline
   Teacher/student & KD &PWKD & CLR & Top1~($\Delta$) \\
   \hline\hline
   None/ResNet-20~(baseline) &  & & & 68.15~(+0.00)\\
   \hline
   None/ResNet-20 & & & $\surd$ & 69.53~(+1.38) \\
   \hline
   ResNet-20$\times$4/ResNet-20 & $\surd$ & & & 68.62~(+0.47) \\
   \hline
   ResNet-20$\times$4/ResNet-20 & &$\surd$ & & 70.44~(+2.29) \\
   \hline
   ResNet-20$\times$4/ResNet-20 & $\surd$ & & $\surd$ & 68.07~(-0.08) \\
   \hline
   ResNet-20$\times$4/ResNet-20 & &$\surd$ & $\surd$ & \textbf{70.76~(+2.61)}\\
   \hline
   \end{tabular}
   \end{center}
   \vspace{-2em}
   \caption{Ablate effects of PWKD and cyclical learning rate scheduler on knowledge distillation efficacy. } 
   \label{tab.6}
\end{table}
\vspace{-0.5em}

Considering the benefit of CLR, we further investigate more possibility of learning rate decay form in each cycle. Besides the rectangular form presented in \cite{smith2017cyclical}, we instantiate other custom learning rate decay policy, multi-step scheduler, linear scheduler and cosine scheduler into CLR. We report the comparison results of different CLR in Table~\ref{tab.7}. We observe that all four CLRs can gain with at least $1.48\%$ improvement. CLR with the rectangular form works best with PWKD and improve ResNet-44 and ResNet-56 with $2.37\%$ and $2.23\%$ respectively.
\begin{table}[htbp]
  \scriptsize
  \begin{center}
  \begin{tabular}{l|c|c}
  \hline
  Method & ResNet-44$\times$4/ResNet-44 & ResNet-56$\times$4/ResNet-56 \\
  \hline\hline
  Vanilla student & 71.17~(+0.00) & 72.07~(+0.00)\\
  \hline
  PWKD~(multi-step) & 72.65~(+1.48) & 73.55~(+1.48) \\
  \hline
  PWKD~(cosine)    & 72.89~(+1.72) & 73.92~(+1.85) \\
  \hline
  PWKD~(linear)    & 73.25~(+2.08) & 74.14~(+2.07) \\
  \hline
  PWKD~(rectangle) & \textbf{73.54~(+2.37)}  & \textbf{74.30~(+2.23)}\\
  \hline
  \end{tabular}
  \end{center}
  \vspace{-2em}
  \caption{Ablate effects of different cyclical learning rate scheduler on the efficacy of PWKD on CIFAR-100.} 
  \label{tab.7}
\end{table}

\subsubsection{Varying groups of knowledge decomposition}
\begin{figure}[htbp]
   \centering
   \begin{minipage}[t]{0.45\linewidth}
       \centering
       \includegraphics[scale=0.25]{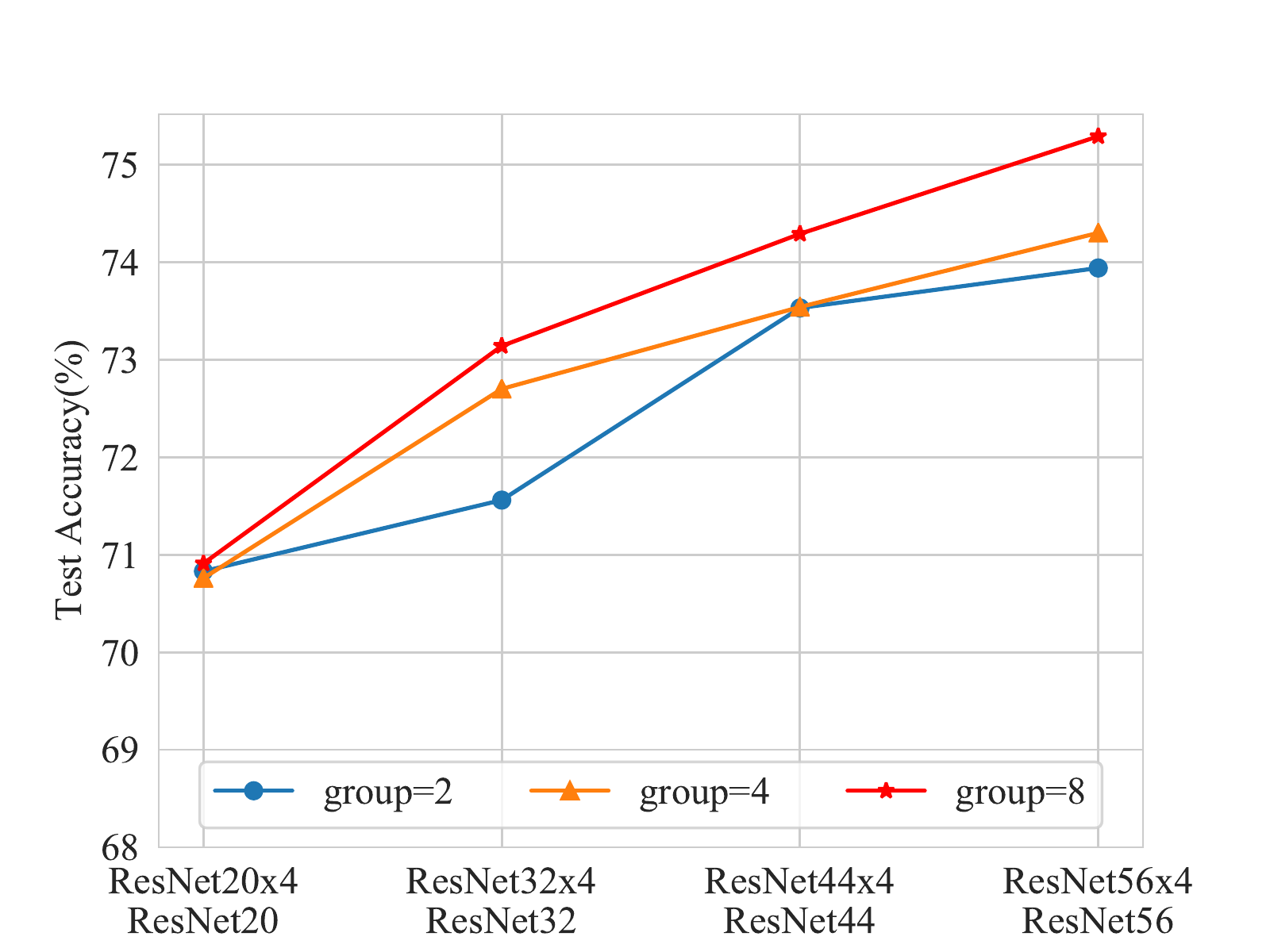}
       \vspace{-2em}
       \caption{Student distills from teacher with varying knowledge group.}
       \label{fig.4}
   \end{minipage}
   \begin{minipage}[t]{0.45\linewidth}
       \centering
       \includegraphics[scale=0.25]{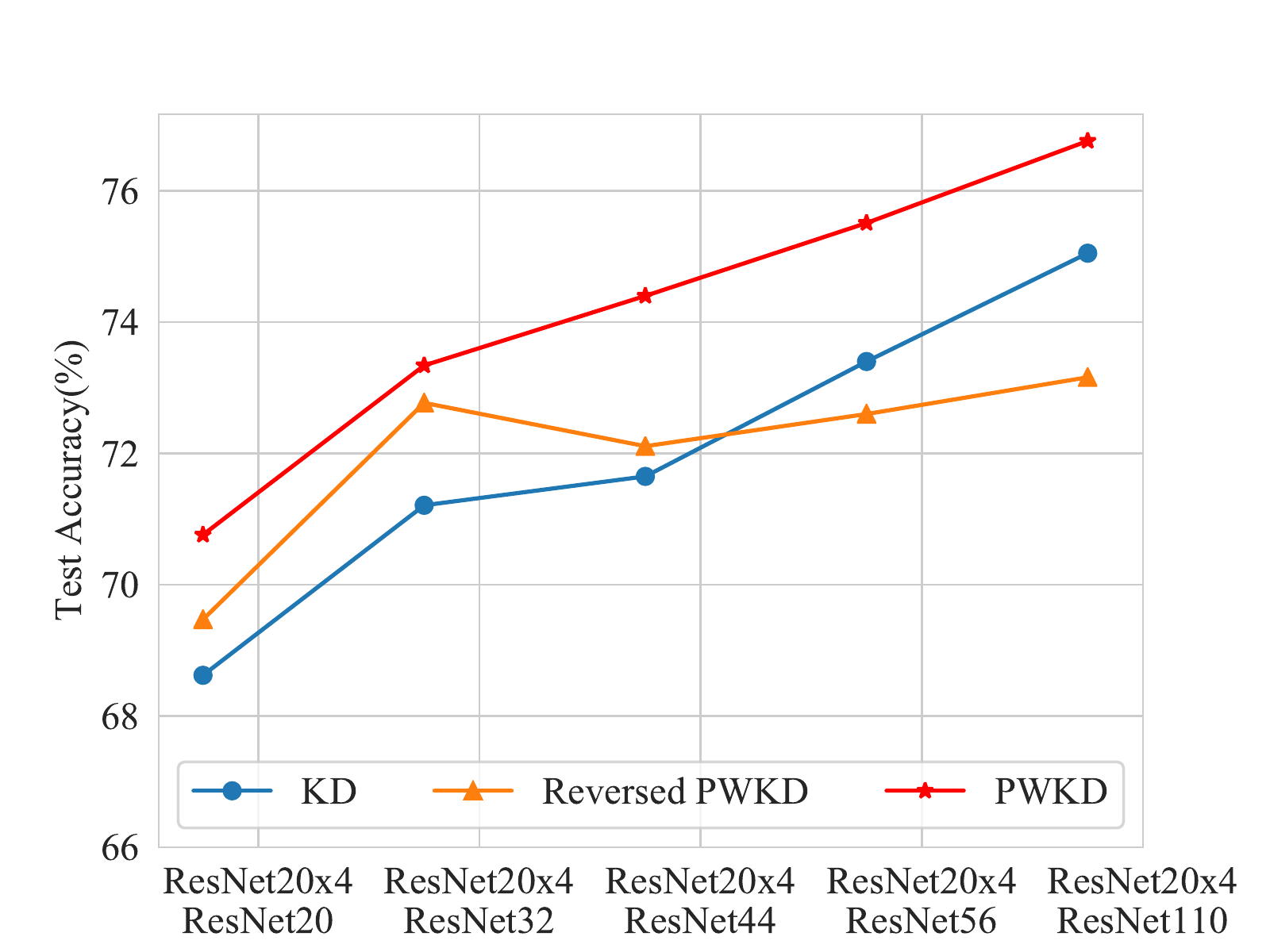}
       \vspace{-2em}
       \caption{Comparisons among vanilla KD, reversed PWKD and PWKD on CIFAR-100.}
       \label{fig.5}
   \end{minipage}
\end{figure}

Knowledge decomposition is one of the key components in PWKD framework, and the groups of knowledge decomposition have great influence on the efficacy of knowledge distillation. Therefore, we investigate the effects of knowledge decomposition by varying groups of knowledge decomposition. Given a teacher, we change the number of sub-networks $G\in\{2,4,8\}$\footnote{Channel width list [0.5$\times$,1.0$\times$], [0.25$\times$,0.5$\times$,0.75$\times$,1.0$\times$] and [0.25$\times$,0.35$\times$,0.45$\times$,0.55$\times$,0.65$\times$,0.75$\times$,0.85$\times$,1.0$\times$] corresponds to G=2,4,8 respectively.}, and train sub-networks jointly to obtain decomposed knowledge. As Figure~\ref{fig.4} shows, given four teacher-student pairs, the test accuracy of student is positive correlated with G. These results seems make sense, and further support the intuition of this paper that smoother distillation boosts student better. However, considering the trade-off between teacher training computation and student distillation performance, we set G to 4 in default.

\subsubsection{Varying knowledge distillation order}
As claimed in the introduction, curriculum learning in distillation process is such significant, and we further investigate the knowledge distillation order in this section. We consider two types of distillation order in PWKD, that is curriculum order~(distilling knowledge from $\Phi_{0.25\times}\rightarrow\Phi_{0.5\times}\rightarrow\Phi_{0.75\times}\rightarrow\Phi_{1.0\times}$) and reversed order~(distilling knowledge from $\Phi_{1.0\times}\rightarrow\Phi_{0.75\times}\rightarrow\Phi_{0.5\times}\rightarrow\Phi_{0.25\times}$). We set ResNet-20$\times$4 as teacher, and ResNet-20, ResNet-32, ResNet-44, ResNet-56 and ResNet-110 as student. We compare PWKD and reversed PWKD in Figure~\ref{fig.5}, and results show that PWKD consistently outperform reversed PWKD across five teacher-student pairs. In some cases~(\textit{e.g.}, ResNet-20$\times$4/ResNet-56 and ResNet-20$\times$4/ResNet-110), reversed PWKD even has inferior performance than vanilla KD~\cite{hinton2015distilling}. We conclude that curriculum learning from partial to whole is such promising for knowledge distillation.




\section{Conclusion}
We analyze knowledge distillation from a new perspective of teacher knowledge quantity and propose PWKD paradigm. Comprehensive evaluation results across datasets, distillation approaches and teacher-student pairs demonstrate the generality and effectiveness of PWKD. We further conclude that partial-whole hypothesis actually makes sense and knowledge quantity has the potential to improve the efficacy of knowledge distillation. In the future, we plan to explore more knowledge decomposition paradigms from the angle of feature resolution, network depth and etc. Relating knowledge distillation and adaptive neural networks is an interesting topic.
\clearpage
{
\bibliography{reference.bib}
}

\end{document}